\pdfoutput=1
\documentclass[11pt]{article}
\usepackage{acl}
\usepackage{booktabs}
\usepackage{graphicx}
\usepackage{tabularx}
\usepackage{xspace}
\usepackage{dirtytalk}
\usepackage{times}
\usepackage{latexsym}
\usepackage[T1]{fontenc}
\usepackage[utf8]{inputenc}
\usepackage{microtype}
\usepackage{inconsolata}
\usepackage{amsmath}
\usepackage{amssymb}
\usepackage{amsfonts}
\usepackage{multirow}
\usepackage{scalefnt}
\usepackage{paralist}
\usepackage[nameinlink]{cleveref}
\renewcommand{\autoref}[1]{\Cref{#1}}

\newcommand{\emotionname}[1]{\textit{#1}}
\newcommand{\fear}{\emotionname{fear}\xspace}
\newcommand{\joy}{\emotionname{joy}\xspace}
\newcommand{\anger}{\emotionname{anger}\xspace}

\newcommand{\trust}{\emotionname{trust}\xspace}

\newcommand{\guilt}{\emotionname{guilt}\xspace}
\newcommand{\shame}{\emotionname{shame}\xspace}

\newcommand{\surprise}{\emotionname{surprise}\xspace}
\newcommand{\sadness}{\emotionname{sadness}\xspace}
\newcommand{\anticipation}{\emotionname{anticipation}\xspace}
\newcommand{\disgust}{\emotionname{disgust}\xspace}

\newcommand{\other}{\emotionname{other}\xspace}
\newcommand{\noemotion}{\emotionname{no emotion}\xspace}

\newcommand{\pleasantness}{\emotionname{pleasantness}\xspace}
\newcommand{\pride}{\emotionname{pride}\xspace}
\newcommand{\effort}{\emotionname{effort}\xspace}

\newcommand{\certainty}{\emotionname{certainty}\xspace}

\newcommand{\selfResp}{\emotionname{self responsibility}\xspace}
\newcommand{\responsibility}{\emotionname{responsibility}\xspace}

\newcommand{\selfControl}{\emotionname{self control}\xspace}

\newcommand{\control}{\emotionname{control}\xspace}

\newcommand{\corpusname}[1]{{\small \textsc{#1}}}

\newcommand{\isear}{\corpusname{ISEAR}\xspace}
\newcommand{\ssec}{\corpusname{SSEC}\xspace}
\newcommand{\tales}{\corpusname{Tales}\xspace}

\newcommand{\appreddit}{\scalebox{0.8}[1]{\corpusname{APPReddit}}\xspace}
\newcommand{\enisear}{\corpusname{enISEAR}\xspace}
\newcommand{\crowdenv}{\corpusname{enVENT}\xspace}
\newcommand{\crowdenvlong}{\corpusname{crowd-enVENT}\xspace}

\newcommand{\bertopic}{\corpusname{BERTopic}\xspace}
\newcommand{\intopic}{\corpusname{InTopic}\xspace}
\newcommand{\crosstopic}{\corpusname{CrossTopic}\xspace}

\newcommand{\ingrl}{\corpusname{InTopic-gr}\xspace}
\newcommand{\crossgrl}{\corpusname{CrossTopic-gr}\xspace}

\newcommand{\bert}{\corpusname{BERT}\xspace}
\newcommand{\roberta}{\corpusname{RoBERTa}\xspace}

\newcommand{\F}{F$_1$\xspace}
\makeatletter
\renewcommand\paragraph{\@startsection{paragraph}{4}{\z@}%
  {0.5ex \@plus1ex \@minus.2ex}%
  {-1em}%
  {\normalfont\normalsize\bfseries}}
\makeatother

\title{Topic Bias in Emotion Classification}

\author{Maximilian Wegge$^1$ \and Roman Klinger$^{1,2}$ \\
  $^1$Institut f\"ur Maschinelle Sprachverarbeitung, University of
  Stuttgart, Germany \\
  $^2$Fundamentals of Natural Language Processing, University of Bamberg, Germany \\
  \texttt{maximilian.wegge@ims.uni-stuttgart.de}\\
  \texttt{roman.klinger@uni-bamberg.de}}

\begin{document}
\maketitle
\begin{abstract}
  Emotion corpora are typically sampled based on keyword/hashtag
  search or by asking study participants to generate textual
  instances. In any case, these corpora are not uniform samples
  representing the entirety of a domain. We hypothesize that this
  practice of data acquisition leads to unrealistic correlations
  between overrepresented topics in these corpora that harm the
  generalizability of models. Such topic bias could lead to wrong
  predictions for instances like ``I organized the service for my
  aunt's funeral.'' when funeral events are over-represented for
  instances labeled with sadness, despite the emotion of pride being
  more appropriate here.
  In this paper, we study this topic bias both from the data and the
  modeling perspective. We first label a set of emotion corpora
  automatically via topic modeling and show that emotions in fact
  correlate with specific topics.  Further, we see that emotion
  classifiers are confounded by such topics. Finally, we show that the
  established debiasing method of adversarial correction via gradient
  reversal mitigates the issue.
  Our work points out issues with existing emotion corpora and that
  more representative resources are required for fair evaluation of
  models predicting affective concepts from text.
\end{abstract}

\section{Introduction}
\label{sec:intro}
Emotion analysis is typically formulated as the task of emotion
classification, i.e., assigning emotions to textual units such as news
headlines, social media or blog posts.
Emotion classification is applied across various domains, ranging from
political debates \citep{mohammad2014} to dialogs \citep{li2017} and
literary texts \citep{mohammad2011}, and enable further use cases such
as analyzing emotions of social media users (e.g., in response to the
COVID-19 pandemic, \citealp{zhan-etal-2022-feel}), identifying abusive
language using emotional cues \citep{safi-samghabadi-etal-2020} or
developing empathetic dialog agents, e.g., for emotional support
\citep{liu-etal-2021}.

Emotions are thereby modeled as either discrete classes of basic
emotions \cite{ekman1992,plutchik2001}, within the vector space of
valence and arousal \citep{russell1980}, or as the result of the
emoter's cognitive appraisal of the stimulus event
\cite{scherer2005,smith1990}.
Independent of which emotion theory is adopted, emotion data sets are commonly
collected by searching for topics of interest, for instance
with hashtags on social media \citep[i.a.]{Schuff2017} or by using
specific subfora \citep{Stranisci2022}, in order to cover a variety of emotion labels
instead of generally overrepresented ones.
Another common approach
is to ask study participants to report emotional episodes for a given
emotion
\citep[i.a.]{troiano-etal2023,Troiano2019,scherer-wallbott1994}. In
that case, subjects are more likely to report important, long
enduring, high-impact events than less relevant ones. Cases in which
large corpora are uniformly sampled for annotation are comparably rare
\citep[i.a.]{Alm2005}.

We hypothesize that these established sampling procedures are
harmful. They lead to topics overrepresented for specific emotions
which allows the model to rely on spurious signals instead of actual
emotion expressions. As an example, in \emph{``I enjoyed my birthday
  party.''} a model might learn to associate the topic of ``party''
with \joy, instead of inferring the emotion from the text (here, the
verb). That might then lead to wrong predictions for texts such as
\emph{``I did not like my party.''}. We assume that this is also
a reason for poor cross-corpus generalization of emotion
classification \citep[cf.][]{Bostan2018}.

In this paper, we aim at understanding the prevalence and impact of
this phenomenon in the context of emotion analysis.  We answer the
following research questions:
\begin{compactenum}
\item \textit{Are emotion datasets biased towards topics?}\\
  We show that emotion datasets are biased towards topics, i.e., that
  there is a prototypical association of topics with emotion labels
  specific for each corpus.
\item \textit{Is emotion classification influenced by topics?}\\
  Based on the observation of topic biases in datasets, we show that this
  bias also carries over to emotion prediction models.
\item \textit{Can the influence of topics on emotion classification be mitigated?}\\
  We show that the robustness of emotion classifiers can be improved
  by using established debiasing methods which reduce the impact of
  the topic bias on the classifiers.
\end{compactenum}
We perform the experiments on emotion self-report corpora
\citep{scherer-wallbott1994, troiano-etal2023, Hofmann2020}, social media data from
Twitter \citep{Schuff2017} and Reddit \citep{Stranisci2022}, as well
as on fictional stories \citep{Alm2005}. With these annotated corpora,
we cover (i) a variety of domains and (ii) multiple emotion models.

\section{Related Work} \label{sec:rel_work}
\subsection{Emotion Classification} \label{subsec:emo_class}
Computational approaches to emotion analysis often adopt categories
inspired by theories of basic emotions \cite{ekman1999,plutchik1982}, by modeling
emotions as six (\anger, \fear, \joy, \sadness, \disgust, \surprise)
or eight (adding \anticipation, \trust) discrete classes. Alternatives
include the use of the valence--arousal vector space to position
emotion categories \citep{russell1980} or focus on the aspect that
emotions are caused by events that undergo a cognitive evaluation
\citep{scherer2005,smith1990}. In the latter case, emotions are
represented by appraisal variables, including, for instance, if the
event requires attention, if the person involved is certain about what
is happening, if the outcome requires further effort, is pleasant, or
if the person has been responsible or can control the situation.

The emotion model is sometimes, but not always, chosen based on the
domain a corpus stems from. For instance, \newcite{Schuff2017}
reannotate a stance detection corpus with Plutchik's eight emotions
due to their presumed universality. \newcite{Alm2005} follow Ekman's
model for a similar reason. \newcite{scherer-wallbott1994,Hofmann2020}
choose a set of self-directed emotions because their data consists of
self-reports. \newcite{troiano-etal2023} use a larger set of emotions,
and also annotate appraisal dimensions because of the prevalence of
event descriptions in the texts they collected, similarly to
\newcite{Stranisci2022}.

To develop automatic emotion classification methods, as in many areas
of NLP, transformer-based pre-trained language models like \bert
\citep{devlin-etal-2019-bert} and \roberta \citep{liu2019roberta} have
been found to consistently outperform previous state-of-the-art
approaches. These models are fine-tuned on domain-specific
corpora. \citet{Bostan2018} show for 14 popular emotion datasets that
a cross-corpus prediction performance is drastically lower than for
in-corpus classification.  We hypothesize that a major part of what
makes a domain unique is the distribution of topics.

\subsection{Bias} \label{subsec:bias}
Bias has been found to affect various textual resources, including
those to support hate-speech detection \citep{wich2020}, sentiment
analysis \citep{wang2021}, machine translation \citep{stanovsky2019}
or argument mining \citep{spliethover-wachsmuth-2020-argument}.  In
general, the term bias refers to the phenomenon that machine
learning models adopt latent, \say{non-generalizable features}
\citep{shah-etal-2020-predictive} from the training data, such as
domain-specific terms, contexts, or text styles.  In consequence, the
biased representation leads to erroneous results when applied to a
domain where the alleged standard does not hold
\citep[cf.][]{hovy2021-bias}, which can lead to harmful impact on
various groups in our society.

Topic bias originates in skewed topic
representations. \citet{wiegand2019}, for instance, find the topic of
\textit{soccer} to be almost exclusively associated with abusive
language, caused by the sampling procedure.
In this paper, topic bias is understood to comprise two of these
concepts: First, the association of certain emotion or appraisal
labels with certain topics and second, the resulting bias in a
classifier towards certain topics when predicting the emotion and
appraisal labels.

\paragraph{Detection and Mitigation.}
For detecting bias contained within pre-trained models and word
embeddings, \citet{caliskan2017} introduce the Word Embedding
Association Test (WEAT) and \citet{kurita2019} investigate gender bias
within \bert word embeddings. \citet{wiegand2019} calculate the
pointwise mutual information between words and abusive language
annotations.  \citet{nejadgholi-kiritchenko-2020-cross} train a topic
model on a dataset and perform a qualitative analysis of the result.

Bias mitigation is addressed at either the data or the modeling
level. \citet{wiegand2019} sample additional texts of the
overrepresented class. \citet{barikeri2021} augment training data by
instance duplication, replacing the biased term with an inverse term.
\citet{he2019} tackle the bias correction during training by
developing an intentionally biased classifier in order to identify the
features that exhibit bias. This information is then used to train a
debiased classifier which compensates for the biased features.
\citet{qian2019} adapt the language model's loss function in order to
mitigate gender bias, introducing a new term to the loss function that
aims at equalizing the probability of male and female words.  In the
context of mitigating the influence of domains on classification,
gradient reversal has proven effective \citep{ganin2015}.

\section{Methods \& Experimental Setting}
We will now explain our method for topic-bias detection in emotion
corpora and then the experimental setting to evaluate established
mitigation methods in this domain.\footnote{The repository to
  replicate our experiments will be made available via
  \url{https://www.bamnlp.de/resources/}.}
\paragraph{Definitions.} We consider six different corpora, where each
corpus $c \in C$ is modeled as a tuple consisting of a set of topic
labels $T_c$, a set of instances $I_c$ and a set of annotation labels
$L_c$, where $L_c$ is either from the set of overall
appraisals ($L_c \subseteq A_C$) or emotion labels
($L_c \subseteq E_C$), where $A_C \cap E_C = \varnothing$.

Further, each instance $i_c \in I_c$ consists of a text
$s_{i,c} = (s_1, s_2, \dots, s_n)$, a topic label $t_{i,c} \in T_c$ and
a set of emotion or appraisal labels
$L_{i,c} = \{a_j, \dots, a_k\} \subseteq L_c$.
Some of the corpora we consider are labeled with multiple, i.e., one
or more emotions. Appraisals are always annotated in a multi-label
setting.

\begin{table*} % table 7
  \centering\small
  \setlength{\tabcolsep}{4pt}
  \begin{tabularx}{\textwidth}{l r r r X r r}
    \toprule
    & \# Topics & $\varnothing$ Topic & \textsc{StD} & Topic labels & \# Instances & \textit{Outlier} \\
    \cmidrule(r){1-1}
    \cmidrule(lr){2-2}
    \cmidrule(lr){3-3}
    \cmidrule(lr){4-4}
    \cmidrule(lr){5-5}
    \cmidrule(lr){6-6}
    \cmidrule(l){7-7}
    \isear
    & $10$ & $525$ & $290$ & love, exams, death, shame, school, animals, alcohol, accidents, fear, theft & 7666 & 2412\\
    \ssec
    & $11$ & $305$ & $219$ & feminism, prayer, abortion, climate, clinton, twitter, trump, gay marriage, latino, swearing, patriotism & 4870 & 1513\\
    \tales
     & $10$ & $388$ & $183$ & birds, flowers, tabitha twitchit, old english, piggies, royalty, dressmaking, hansel \& gretel, boats, predators & 10339 & 6457\\
    \crowdenv
    & $8$ & $584$ & $298$ & feelings, promotion, relationships, covid, dogs, graduation, pregnancy, driving & 6600 & 1925\\
    \appreddit
    & $10$ & $43$ & $12$ & depression, everyday life, driving, love, romantic relationships, reddit, anger, death, platonic relationships, vaccination & 780 & 352\\
    \enisear
    & $13$ & $58$ & $25$ & death, dogs, accidents, theft, birth, food, affairs, UK politics, christmas, bullying, work, relationships, spooky & 1001 & 245\\
    \bottomrule
  \end{tabularx}
  \caption{Number (\#), average size ($\varnothing$), standard
    deviation (\textsc{StD}), and manually assigned labels of the
    topics found by \textsc{BerTopic} for all corpora. All numbers
    exclude the outlier topic, whose number of instances is provided
    in the last column (\textit{Outlier}). The topic labels are sorted
    by size, in decreasing order. The second to last column reports
    the number of all instances per corpus for reference. (We
    abbreviate the \crowdenvlong corpus in this paper as \crowdenv.)}
  \label{tab:topic_distr_all}
\end{table*}

\subsection{Topic-based Bias Detection}
Inspired by \citet{wiegand2019,nejadgholi-kiritchenko-2020-cross}, we
train separate emotion classifiers tasked with predicting either the
emotion or appraisal label $a \in L_{i,c}$, for each topic
$t^{\mathrm{out}}_c \in T_c$ in a given corpus.  In the subset of the
corpus used for training the classifier ($T^{\mathrm{train}}$),
instances with the topic label $t^{\mathrm{out}}_c$ are excluded,
i.e.,
$T^{\mathrm{train}}_{c} = \{t_{i,c} | t_{i,c} \in T_c, t_{i,c} \neq
t^{\mathrm{out}}_c\}$.
The number of classifiers trained for a given corpus $c$ is thus equal
to $|T_c|$.

\begin{figure}[t]
  \centering
  \includegraphics[width=\linewidth]{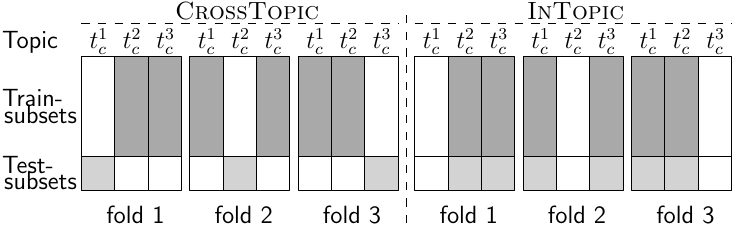}
  \caption{Visualization of the experimental setting for \intopic and \crosstopic
    predictions.}
  \label{fig:expsetup}
\end{figure}

The classifiers are evaluated in two distinct settings: In the
\intopic setting, multiple testsets are sampled from the corpus, one
for each topic except $t^{\textrm{out}}_{c}$.  Each testset is thus
defined in relation to the respective held-out topic:
$t^{\textrm{in}}_c = T_c \setminus \{t^{\textrm{out}}_c\}$.  Thus, the union of
all $t^{\textrm{in}}_c$ per corpus reflects $T^{\textrm{train}}_{c}$.
Therefore, a classifier trained on $T^{\textrm{train}}_c$ is evaluated
on all $t^{\textrm{in}}_c$ of corpus $c$.
For the \crosstopic setting, the classifier is evaluated on the
held-out topic $t^{\mathrm{out}}_c$ which is not part of the training
set $T^{\mathrm{train}}_c$. In both settings, we calculate averages
across folds which leads to a performance estimate whose comparisons
are meaningful. Figure~\ref{fig:expsetup} visualizes this setup.

\paragraph{Topic Modeling.}
While emotion and appraisal annotations stem from the labels of the
respective corpora, the topic labels need to be inferred from the
data. We use \bertopic \citep{grootendorst2022bertopic}, as it
supports pre-trained transformer models to detect the semantic
relations on sentence-level as well as \textsc{HdbScan} for
clustering, averting the need of determining a fixed number of topics
per dataset.  This method has proven effective in previous research
\cite{xu-etal-2022-understanding-narratives,kellert-mahmud-uz-zaman-2022-using,
  eklund-forsman-2022-topic}.

\subsection{Bias Mitigation}
We compare two established methods for debiasing the models with
respect to topics.

\paragraph{Word Removal.}
As a straight-forward approach which still often shows a good
performance \citep[i.a.]{Dayanik2021}, the respective topic words are
removed from the corpus.  Specifically, we remove the most indicative
words for each topic, according to the probabilities of the topic
model.

\paragraph{Gradient Reversal.}
We compare this approach to the well-established method of adversarial
learning through gradient reversal \citep{ganin2015}.  We extend the
emotion/appraisal classifier by a topic predictor and gradient
reversal layer, with the purpose of reversing the gradient (by
multiplying it with $-\lambda$) of the following layer during
backpropagation.
Implementation details for all applied methods are provided in
\autoref{app:implementation}.

\subsection{Data}
We consider six corpora, each annotated for emotions or appraisal
dimensions. We use the \isear \citep{scherer-wallbott1994},
\ssec \citep[Stance Sentiment Emotion Corpus;][]{Schuff2017} and
\tales \citep{Alm2005} corpora for emotion analysis and the \appreddit
corpus \citep{Stranisci2022} for appraisal analysis.  From the
\crowdenv \citep{troiano-etal2023} and \enisear \citep{Troiano2019}
corpora we use both annotation layers.

The corpora differ in size, annotation setup and --
most relevant for us -- in the way the instances are sampled and which
topics are covered:
\isear and \enisear were created by asking study participants to
report and describe events that caused a predefined emotion. \isear
has been collected in an in-lab setup and \enisear via
crowdsourcing. Since participants were free to report any event that
elicited one of the given emotions, they were also free in their
choice of topic. This procedure is in fact expected to create a topic
bias, because more important topics cause more intense emotions and
are therefore more likely to be recalled. Therefore,
\citet{Troiano2019} add diversification method to the otherwise
similar setup. They mention topics that the study participants
shall not report on.

% ssec
In the \ssec corpus, \citet{Schuff2017} re-annotate Twitter posts
originally collected by \citet{mohammad-etal-2016-semeval}. The
original purpose of the text collection was to study sentiment and
stance. Therefore, they have been collected with specific hashtags
corresponding to topics ``Atheism'', ``Climate Change is a Real
Concern'', ``Feminist Movement'', ``Hillary Clinton'', and
``Legalization of Abortion''. Arguably, we could have relied on these
topics in the data, however for comparability in our experiments, we
also use the topic modelling approach for this dataset.

% appreddit
The \appreddit corpus provides appraisal annotations of
Reddit posts, sourced from subreddits mostly connotated with negative
sentiment (Anger, offmychest, helpmecope anxiety, i.a.).
%
% tales
The \tales corpus \citep{Alm2005} features literary texts,
specifically fairy tales by various authors. Here, sentences from
uniformly sampled stories are the unit of annotation.

In order to enable inter-comparability, we map the varying annotation
schemes onto a unified scheme.  More information on the datasets is in
\autoref{app:data}.

\section{Results}
\label{sec:results}
We will now present the results to answer the research questions
introduced in \autoref{sec:intro}.

\subsection{Are emotions biased towards topics?}

\paragraph{Topic Modelling Results.}\autoref{tab:topic_distr_all}
reports the results of the topic modeling at the overall corpus level,
including the number of topics, the average size (number of instances)
and the list of topic labels ($L_c$) for each corpus.
The topic labels are defined manually, based on the ten most
representative words for each topic.

\begin{figure}[t]
  \centering
  \includegraphics[scale=0.30]{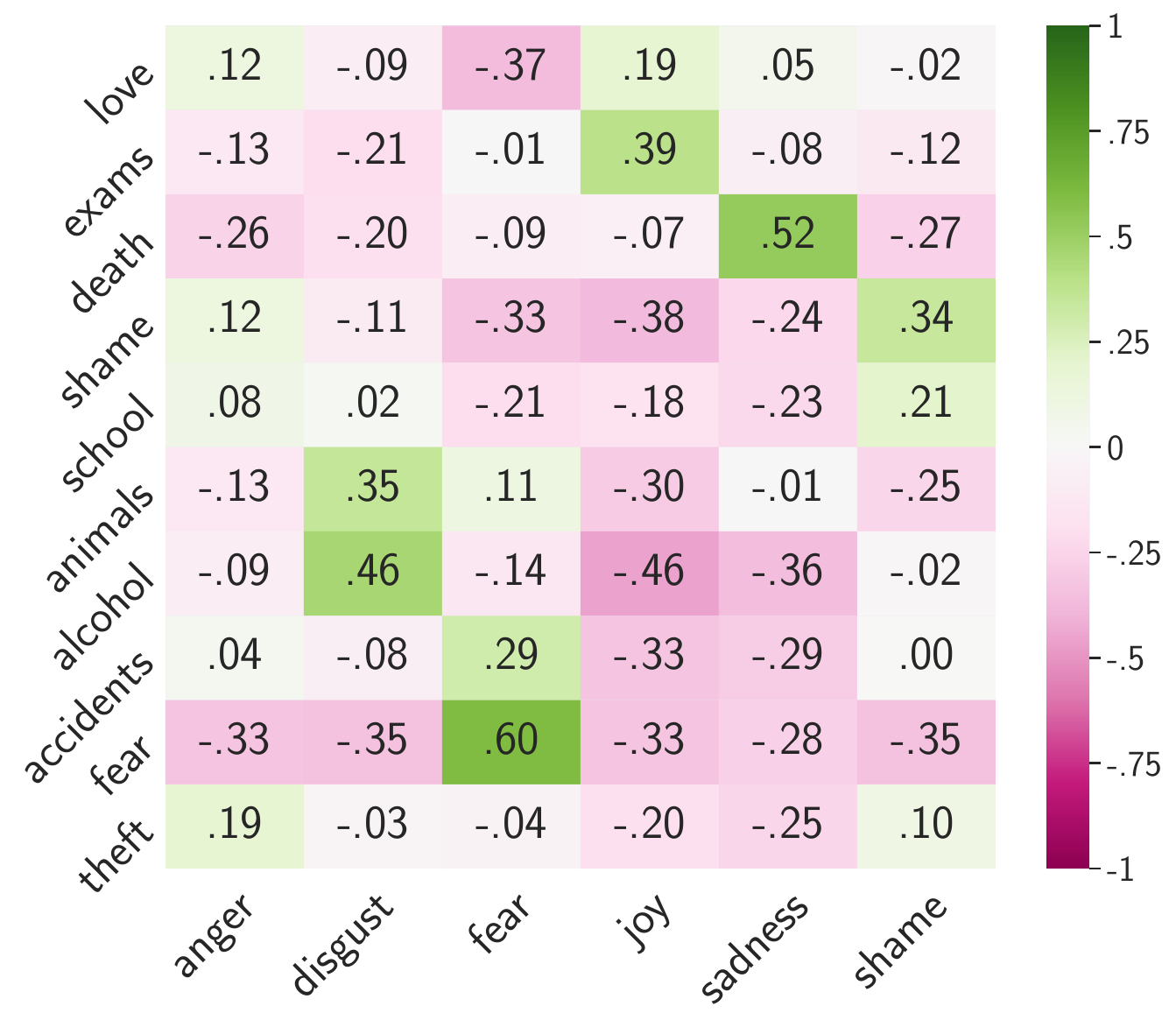}
  \caption{Normalized pointwise mutual information between topics and emotion annotations in \isear.}
  \label{fig:pmi_isear}
\end{figure}

The size of topics, i.e., the number of instances associated with it,
varies across corpora (see
$\varnothing$ and \textsc{StD}). The number of topics ranges from 8
(\crowdenv) to 13 (\enisear), while \isear, \tales and \appreddit
comprise 10, \isear 11 topics.

An important finding is that, despite not being informed in a
supervised manner regarding the emotion labels, the topics reflect the
individual corpus' domain and sampling methods.  \isear, \enisear and
\crowdenv, all of which are compiled by querying emotionally
connotated event-descriptions, feature generic and everyday topics,
e.g., \textit{love}, \textit{dogs} or \textit{driving}.
In \ssec the topic modeling corresponds to the keyword-based sampling
based on the original intention to perform stance detection.
In \appreddit, topics appear to be indicative of the subreddit they
are sourced from.  For instance, the topic of \textit{depression} is
related to the subreddit \say{mentalhealth}. The variety of
relationship-related topics (\textit{romantic relationships},
\textit{love}, \textit{platonic relationships}) reflects the
various subreddits revolving around these topics, e.g.,
\say{relationship advice} or \say{Dear Ex} (cf.\
\citealp{Stranisci2022} for the exhaustive list of sampled
subreddits).
The topics in \tales appear most varied. Some topics correspond to
generic concepts within fairytales (\textit{birds}, \textit{flowers},
\textit{royalty}), while others are representative of specific fairy
tales\footnote{The most representative terms for the topic labeled
	as \textit{Tabitha Twitchit} comprise the names of fictional
	characters from the kids stories by Beatrix Potter. Further, the
	topic \textit{old english} appears to be based on lexical features alone (e.g., \say{thou}, \say{thee}, \say{thy}).}.

\begin{figure}[t]
  \centering
  \includegraphics[scale=0.3]{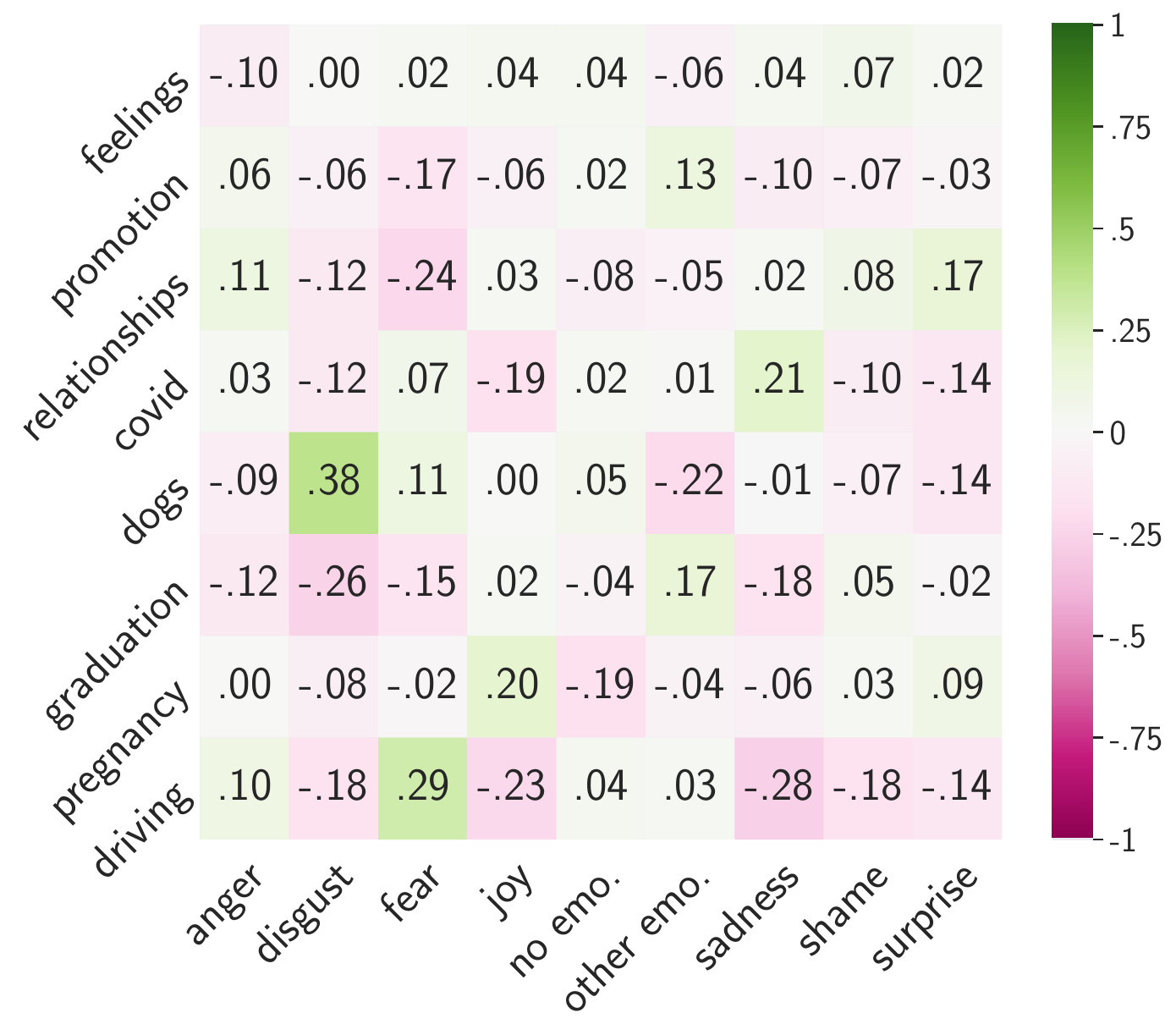}
  \caption{Normalized pointwise mutual information between topics and emotion annotations in \crowdenv.}
  \label{fig:pmi_crowd-env-emo}
\end{figure}
\begin{figure}[t]
  \centering
  \includegraphics[scale=0.3]{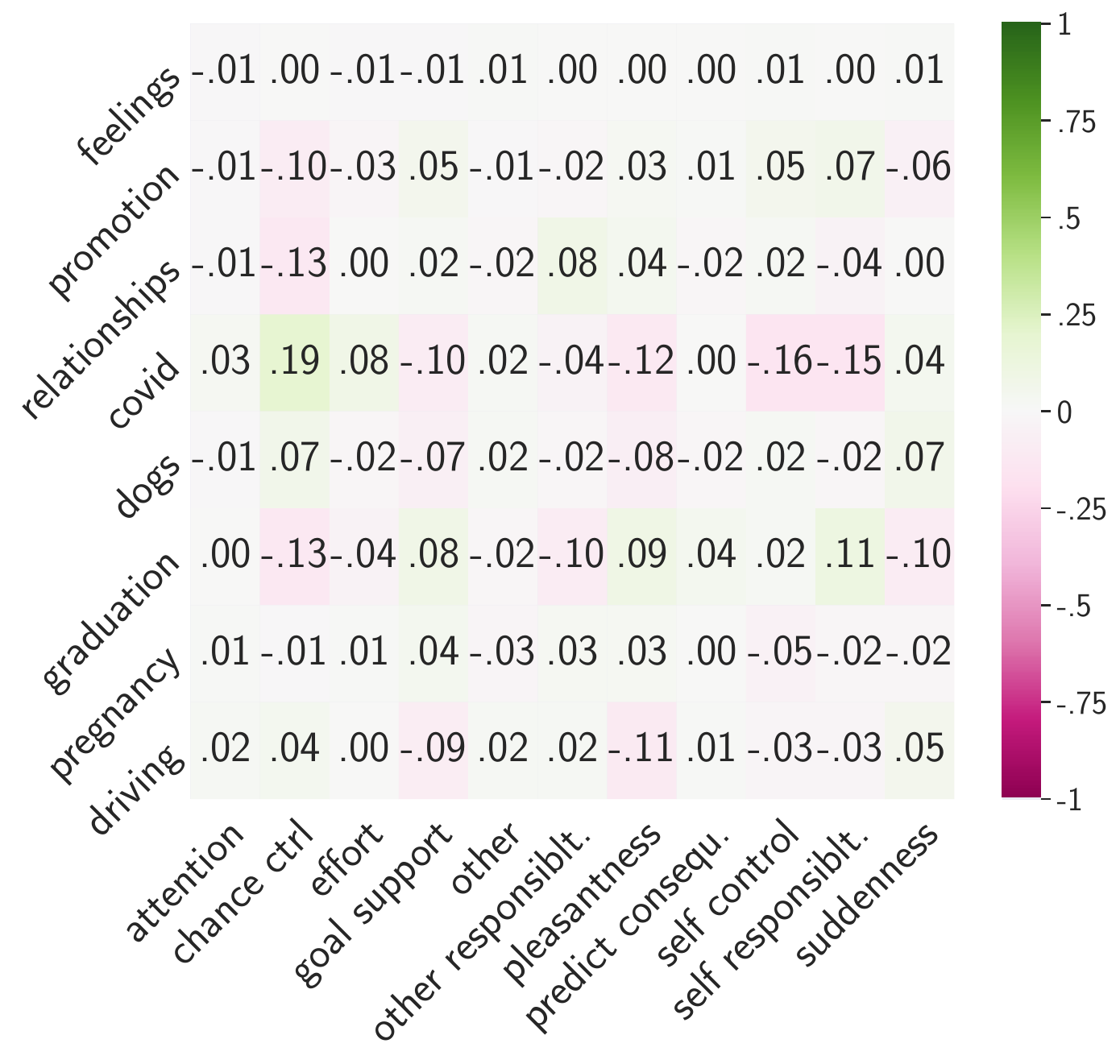}
  \caption{Normalized pointwise mutual information between topics and appraisal annotations in \crowdenv.}
  \label{fig:pmi_crowdenv-appr}
\end{figure}

\begin{table*}[t]
  \centering\small
  \newcommand{\septwo}{\cmidrule(l){2-2}\cmidrule(l){3-3}\cmidrule(l){4-4}\cmidrule(l){5-5}\cmidrule(l){6-6}\cmidrule(l){7-7}\cmidrule(l){8-8}\cmidrule(l){9-9}\cmidrule(l){10-10}\cmidrule(l){11-11}\cmidrule(l){12-12}\cmidrule(l){13-13}\cmidrule(l){14-14}\cmidrule(l){15-15}}
\newcommand{\septhree}{\cmidrule(r){3-3}\cmidrule(r){4-4}\cmidrule(r){5-5}\cmidrule(r){6-6}\cmidrule(r){7-7}\cmidrule(r){8-8}\cmidrule(l){9-9}\cmidrule(l){10-10}\cmidrule(l){11-11}\cmidrule(l){12-12}\cmidrule(l){13-13}\cmidrule(l){14-14}\cmidrule(l){15-15}}
\newcommand{\sepone}{\cmidrule(r){1-2}\cmidrule(l){3-3}\cmidrule(r){4-4}\cmidrule(r){5-5}\cmidrule(r){6-6}\cmidrule(r){7-7}\cmidrule(r){8-8}\cmidrule(l){9-9}\cmidrule(l){10-10}\cmidrule(l){11-11}\cmidrule(l){12-12}\cmidrule(l){13-13}\cmidrule(l){14-14}\cmidrule(l){15-15}}

\newcommand{\phmi}{\phantom{$-1$}}
\setlength{\tabcolsep}{6pt}
\renewcommand{\arraystretch}{1.1}
  \begin{tabular}{ll ccc rr ccc rr rrr }
    \toprule
    & & \multicolumn{5}{c}{\crosstopic} & \multicolumn{5}{c}{\intopic} & \multicolumn{3}{c}{$\Delta^{\textsc{InTopic}}_{\textsc{CrossTopic}}$} \\
    \cmidrule(r){3-7}\cmidrule(r){8-12}\cmidrule(){13-15}
    & Corpus
    & \textsc{Bl}&\textsc{Wr}&\textsc{Gr}&$\Delta^{\textsc{Bl}}_{\textsc{Wr}}$&$\Delta^{\textsc{Bl}}_{\textsc{Gr}}$
    &\textsc{Bl}&\textsc{Wr}&\textsc{Gr}&$\Delta^{\textsc{Bl}}_{\textsc{Wr}}$&$\Delta^{\textsc{Bl}}_{\textsc{Gr}}$
    &\textsc{Bl}&\textsc{Wr}&\textsc{Gr} \\
    \sepone
    \multirow{5}{*}{\rotatebox{90}{Emotion}}
    & \isear    & 59 & 59 & 65 &0    &6     & 68 & 70 & 71 &2   &3     & 9 & 11 & 6 \\
    & \enisear   & 69 & 54 & 68 & $-$15 & $-$1 & 74 & 69 & 72 & $-$5 & $-$2 & 5 & 15 & 4 \\
    & \ssec     & 46 & 37 & 23 &$-$12&$-$23 & 47 & 39 & 25 &$-$8&$-$22 & 1 & 2 & 2 \\
    & \tales    & 84 & 84 & 82 &0    &$-$2  & 85 & 85 & 83 &0   &$-$2  & 1 & 1 & 1 \\
    & \crowdenv & 51 & 51 & 54 &0    &3     & 55 & 55 & 57 &0   &2     & 4 & 4 & 3 \\
    \septwo
    & Average
    & 62 & 57 & 58 &$-$5&$-$4 & 66 & 64 & 62& $-$2 & $-$4 & 4 & 7 & 4 \\
    \midrule
    \multirow{5}{*}{\rotatebox{90}{Appraisal}}
    & \enisear   & 70 & 56 & 54 &$-$14 &$-$16 & 75 & 57 & 56 &$-$18&$-$19 & 5 & 1 & 2 \\
    & \crowdenv  & 63 & 61 & 44 &$-$2  &$-$19 & 64 & 61 & 45 &$-$3 &$-$19 & 1 & 0 & 1 \\
    & \appreddit & 66 & 56 & 56 &$-$10 &$-$10 & 68 & 55 & 56 &$-$13&$-$12 & 2 & $-$1 & 0 \\
    \septwo
    & Average
    & 66 & 57 & 51 &$-$9&$-$15 & 69 & 57 & 52 &$-$12&$-$17 & 3 & 0 & 1 \\
    \bottomrule
  \end{tabular}
  \caption{Results for \crosstopic and \intopic experiments and
    differences between them for all experimental series. For each
    experimental setup, we show results for the baseline without
    debiasing (\textsc{Bl}) and for the two debiasing methods of word
    removal (\textsc{Wr}) and gradient reversal (\textsc{Gr}).}
  \label{tab:in_corpus_all_f1}
\end{table*}

\paragraph{Emotion--Topic Relation.} We will now look at the relation
between emotions and topics from the dataset perspective.  At first
glance, such relations can already be observed in topics that revolve
around specific emotions, such as \textit{shame}, \textit{fear} (both
in \isear), \textit{anger} (\appreddit) or, more general,
\textit{feelings} (\crowdenv).
In order to assess whether these equivalences on the lexical level are
also present in the respective emotion annotations, we report the
normalized pointwise mutual information between topics and their
associated emotion annotations in Figures~\ref{fig:pmi_isear} and
\ref{fig:pmi_crowd-env-emo}.\footnote{We focus our analysis on select
  datasets and report results for the remaining corpora in
  \autoref{app:pmi}.}
For \isear (Fig.~\ref{fig:pmi_isear}), we observe that the topics of
\textit{shame} and \textit{fear} are positively correlated with the
emotion label of the same class.  Further, emotionally correlated
topics are \textit{death} (with \sadness), \textit{alcohol} and
\textit{animals} (both \disgust), \textit{accidents} (\fear) and
\textit{exams} with \joy (all positive).  Negative correlations can be
observed for \textit{alcohol} and \joy, as well as for \textit{love}
and \fear.

The observations for \crowdenv are similar
(Fig.~\ref{fig:pmi_crowd-env-emo}), with positive correlations between
\textit{dogs} and \disgust as well as \textit{driving} and
\fear. Although these are consistent with correlations of similar
topics in \isear (\textit{animals} and \disgust, \textit{accidents}
and \fear), the PMI values in \crowdenv are consistently lower.

The \crowdenv offers itself to compare the emotion--topic and
appraisal--topic correlations
(\autoref{fig:pmi_crowdenv-appr}).
The highest positive correlation is between \textit{covid} and
\textit{chance control}, i.e., \textit{covid}-related events are
appraised as out of control by the emoter. The topic of \textit{covid}
is further (slightly) negatively correlated with \selfControl (thus,
the complement to \textit{chance control}) and \selfResp.  This direct
comparison on \crowdenv shows that the correlations between topics and
appraisals are less distinct than for emotions.

\begin{figure*}[t]
  \begin{minipage}{.49\linewidth}
    \centering
    {\tiny\sf No bias mitigation}\\[-0.1\baselineskip]
    \includegraphics[width=0.9\linewidth]{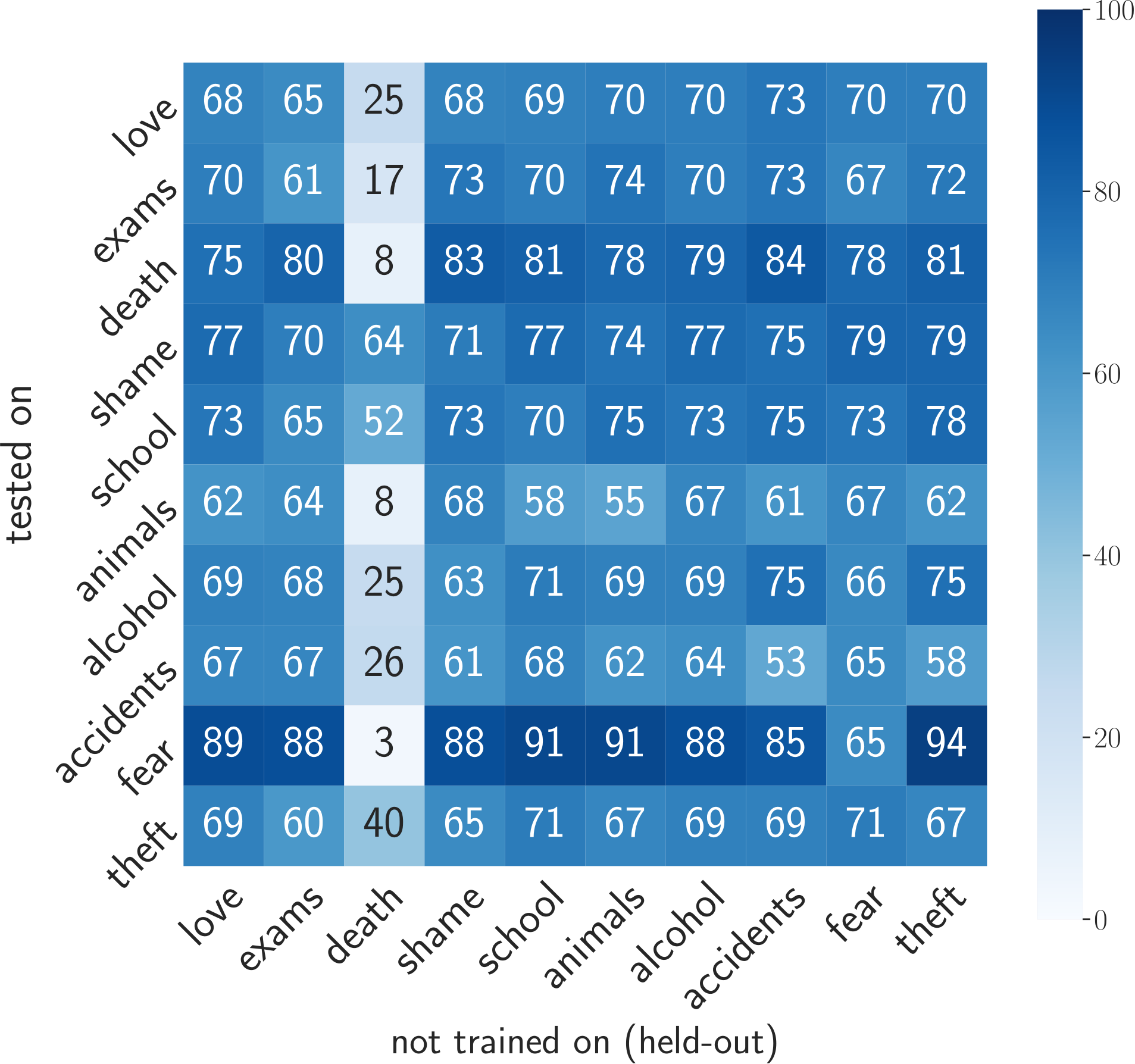}
    \caption{Micro-average \F for each topic-specific test set in \isear, for each held-out topic (\crosstopic/\intopic). No mitigation method is used (\textsc{Bl} setting).}
    \label{fig:micros_base_isear}
  \end{minipage}
  \hfill
  \begin{minipage}{.49\linewidth}
    \centering
    {\tiny\sf gradient reversal mitigation}\\[-0.1\baselineskip]
    \includegraphics[width=0.9\linewidth]{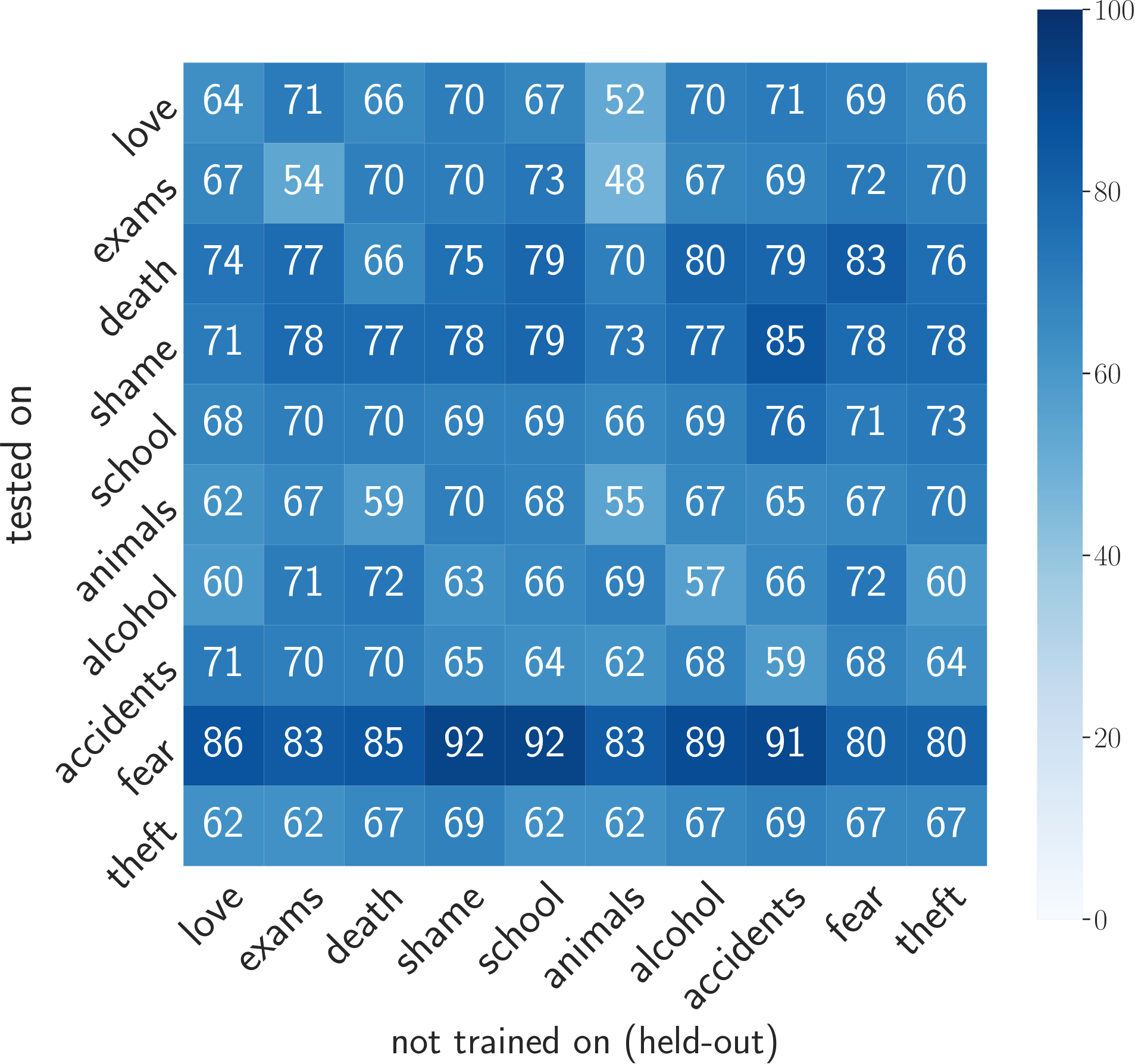}
    \caption{Micro-average \F for each topic-specific test set in \isear, for each held-out topic (\crosstopic/\intopic). Gradient reversal is used as a mitigation method (\textsc{Gr} setting).}
    \label{fig:micros_grl_isear}
  \end{minipage}
\end{figure*}

\subsection{Is emotion classification influenced by topics?}
What arises from the observation that topics and emotions (and topics
and appraisals) are indeed correlated is the question whether this
relation is reflected in classifiers.  To this end,
\autoref{tab:in_corpus_all_f1} shows results for \crosstopic and
\intopic experiments.

Following the assumption that emotion and appraisal classifiers are
biased towards topics, the \intopic setting is hypothesized to score
higher than the \crosstopic setting. The difference between these two
settings is shown in the
$\Delta^{\textsc{InTopic}}_{\textsc{CrossTopic}}$--\textsc{Bl}
column. Across all corpora, we see that all \intopic scores are higher
than the \crosstopic scores -- the $\Delta$ is positive but varies:
The highest discrepancy is observed for \isear (+9), while it is
neglectable for \ssec, \tales and \crowdenv (in the appraisal
classification setting) and \appreddit (+2).  In comparison, \crowdenv
(for emotion classification) as well as emotion and appraisal classification
on \enisear show moderate improvement when evaluated \intopic
(+4, +5, +5, respectively).  Overall, the $\Delta$ values are similar
(on average) between emotion and appraisal classification.

These results show that the topic influences the predictions
negatively, but does not allow any insight if these results mostly
stem from one emotion label or are the same across labels. To analyze
this aspect, \autoref{fig:micros_base_isear} reports the \F-scores
obtained on each topic-specific subset for each held-out topic. The
diagonal thus depicts the \crosstopic setting. All other cells
correspond to the \intopic setting.

The large $\Delta$ value reported for \isear in
Table~\ref{tab:in_corpus_all_f1} leads to the diagonal values
(\crosstopic) in \autoref{fig:micros_base_isear} to be lower than the
average of all other results of the same held-out topic
(\intopic). However, the \crosstopic scores are still comparably high.
Particularly interesting is the topic of \textit{death}. When this is
absent from the training data, the classifier performs much worse on
all testsets, both \intopic and \crosstopic.  Analogously, the topic
\textit{fear} appears to contain instances easier to classify, no
matter which held-out topic is absent from the training data. The only
exception is the mentioned topic \textit{death}, and, although to a
lesser extent, the \crosstopic setting of the topic \fear.

\subsection{Can the influence of topics on emotion classification be
  mitigated?}
To understand if the discrepancy between the \crosstopic and \intopic
results can be mitigated with debiasing methods, we show the results
also in \autoref{tab:in_corpus_all_f1} (columns \textsc{Wr} for word
removal and \textsc{Gr} for gradient reversal).

\textit{Do the mitigation methods lower the performance for each
  setting separately or do they improve it?} The answer can be found
in the $\Delta^{\textsc{Bl}}_{\textsc{Wr}}$
$\Delta^{\textsc{Bl}}_{\textsc{Gr}}$ columns. In the \intopic setting,
most of these values are negative -- the mitigation method removes
information helpful for emotion classification. The only exception is
the \isear corpus for emotion classification, where the method in fact
improves the result. The negative difference is most pronounced for
\ssec and nearly negligible for the other corpora for emotion
classification. The results carry over to the \crosstopic setting: For
\ssec, emotion classification performance is substantially lower,
while the difference is neglectable for most other corpora. Only
\enisear (for emotion classification) shows a similarly significant
drop in performance when \textsc{Wr} is applied. For \isear, however,
the emotion classification is improved. To provide more detail on
where this \crosstopic improvement takes place, we compare the
detailed \intopic/\crosstopic results for the \textsc{Bl}-and
\textsc{Gr}-settings in Figures \ref{fig:micros_base_isear}
and~\ref{fig:micros_grl_isear}, respectively.  The direct comparison
shows that the substantial impact of the topic \textit{death} on
\crosstopic emotion classification (\autoref{fig:micros_base_isear})
is mitigated when applying the \textsc{Gr}-mitigation method
(\ref{fig:micros_grl_isear}).

\textit{Do the mitigation methods lower the performance discrepancy
  between the \intopic and \crosstopic predictions?}  To find the
answer to this question, we compare the delta values
\textsc{Bl}--\textsc{Wr} and \textsc{Bl}--\textsc{Gr} at the right of
Table~\ref{tab:in_corpus_all_f1}
($\Delta^{\textsc{InTopic}}_{\textsc{CrossTopic}}$). A lower delta
value for the mitigation method than for the \textsc{Bl} is an
indicator that the method improves the classifier. In the emotion
classification setup, this is the case for \isear and, to a lower
extend, for \enisear and \crowdenv. These are the corpora that are
particularly designed to include event descriptions. However, there is
a difference in performance between the mitigation methods. In the
aforementioned corpora, an improvement can only be observed in the
\textsc{Gr}-setting. When \textsc{Wr} is applied, \isear and \enisear
even show a decrease in performance.While the \ssec corpus would also
have the potential to be improved with the method, the classifier
relied too substantially on the topic information and cannot find
enough signal for emotion classification such that the method may
work.

For the appraisal prediction, we also observe an improvement for
event-centered corpora \enisear and \appreddit, but not for
\crowdenv. Throughout all experiments, we observe that topic
information removal is disadvantageous for appraisal prediction.  We
take this as an indicator that the classifiers indeed find information
on the emotion expression outside of topic information. However, the
appraisal information needs to be inferred from the topic of the text
and cannot be found elsewhere.

\begin{table*}
  \centering\small
  \begin{tabularx}{1\linewidth}{lXllccc}
    \toprule
    & & & & \multicolumn{3}{c}{\crosstopic}\\
    \cmidrule(l){5-7}
    ID & Text & Topic & Gold & \textsc{Bl} & \textsc{Wr} & \textsc{Gr}\\
    \cmidrule(r){1-1} \cmidrule(r){2-2} \cmidrule(r){3-3} \cmidrule(r){4-4} \cmidrule(lr){5-5} \cmidrule(lr){6-6} \cmidrule(lr){7-7}
    1 & When one of my closest friends died unexpectantly
      & death & sadness & joy & disgust & \textbf{sadness} \\
    2 & When my uncle comes (3 times a year) for the traditional Christmas dinner with my grandparents and other relatives and is very drunk.
      & alcohol & disgust & anger & shame & \textbf{disgust} \\
    3 & When my fiancee travelled 2000 Km to visit me, and I hadn't seen her for 4 months.
      & love & joy & sadness & sadness & \textbf{joy} \\
    4 & Passing an exam I did not expect to pass.
      & exam & joy & fear & fear & fear \\
    5 & When I was admitted to a certain school as a student.
      & exam & joy & shame & shame & \textbf{joy} \\
    6 & Unexpected visit by a close friend, whom I hadn't seen for half a year.
      & love & sadness & sadness & fear & \textbf{sadness}\\
    \bottomrule
  \end{tabularx}
  \caption{Example predictions for instances from the \isear corpus, including assigned topic and gold emotion label. Predictions are reported for the \crosstopic-setting (trained on all instances except those labeled with respective topic in column \textit{Topic}) when applying no mitigation method (\textsc{Bl}), word removal (\textsc{Wr}) and gradient reversal (\textsc{Gr}). Predictions in \textbf{bold} represent correspondence with gold label.}
  \label{tab:analysis_examples}
\end{table*}

\section{Analysis}
To provide an intuition how the predictions of the model changes
with the topic mitigation, we show examples in
Table~\ref{tab:analysis_examples}.
For each example sentence we see the
corresponding topic label (according to the topic model), the gold
emotion annotation and the \crosstopic-predictions with (\textsc{Wr},
\textsc{Gr}) or without (\textsc{Bl}) applying de-biasing methods.

Example 1 is assigned the topic \textit{death} and is annotated with
\sadness. With no mitigation method applied, a \crosstopic-classifier
(i.e., which has not seen any sentences belonging to the topic
\textit{death} during training) falsely predicts \joy (\textsc{Bl}).
We hypothesize that the erroneous classification is due to a bias
towards the topic of \textit{love} (which is correlated with \joy),
represented by the term \say{friends}.  If word removal is applied, a
different but equally incorrect label is predicted
(\disgust). Apparently, removing any words associated to topics from
the input does mitigate the bias observed in the \textsc{Bl}
prediction, but removes too much information.  However, when using
gradient reversal, the bias is mitigated and the correct label
\sadness is predicted.  Similar cases can be observed in Examples 2,
3, 5 and 6.

Example 4 shows a different pattern.  Despite achieving de-biasing in
the above cases, there are also examples where gradient reversal fails
to mitigate the bias and predict the correct emotion label. None of
the two mitigation methods leads to a correct prediction. Instead, all
\crosstopic-classifiers assign \fear. Presumably, this is because of
the phrase ``did not expect'' which expresses a future-directed,
misalignment with the predictability of events. This aspect might in
itself be another possible form of appraisal bias.

\section{Conclusion}
We based our study on the observation that emotion analysis corpora
are commonly sampled based on keywords or following other methods that
are risky to lead to distributions that are not representative for the
entirety of a domain. We contributed a better understanding how far
this issue can be found in emotion corpora and if models fine-tuned on
them rely on such spurious signals.

The analysis of topic distributions in emotion corpora yields that
they are, indeed, biased towards topics. The degree of bias varies:
Some corpora exhibit prototypical topics for certain emotions, while
in others, only weak correlations between topic and emotion
distribution can be observed. We hypothesize this is because of the
respective sampling strategies: If the sampling method is biased,
i.e., if certain topics are over-represented for a given emotion,
topic bias emerges.

In the cases in which topic and emotion distributions are highly
correlated, this topic bias is also found to be reflected in the
resulting classifier.  For mitigating this bias in emotion
classifiers, gradient reversal proved to be useful. It
allows the classifier to make use of available topic information
without relying solely on it for making the classification decision.

Our results suggest that classifiers in which the topic bias is
mitigated may have a higher performance across corpora, yet, this
needs to be evaluated in future work. Further, we assume that
prompt-learning or other few-shot modeling methods might suffer less
from topic biases in corpora. If this is true, this opens a new
research direction of selecting non-bias-inducing instances for
emotion and appraisal classification.

Finally, the difference between topic--emotion and topic--appraisal
correlations requires further analysis.  We hypothesize that this is
because appraisals are more closely related to events than general
emotion labels.

\section*{Acknowledgements}
This research has been conducted in the context of the CEAT project,
KL 2869/1-2, funded by the German Research Foundation (DFG).

\section*{Limitations}
We presented the first study on topics as unwanted confounders for
emotion analysis. We focused on a set of popular corpora, but cannot
make any judgements regarding corpora that we did not study. We are
confident that similar effects can be found in other resources, but
this still needs to be analyzed.

Another limitation is the pragmatic decision that the contextualized
embeddings used by our emotion/appraisal predictors and the topic
modeler are not the same. The representations used for topic
clustering are provided by sentence-transformer models, while we
leverage \roberta embeddings for emotion and appraisal classification.
This potentially introduces an uncontrolled variance in our
experiments.  Using identical embedding models for both steps -- or,
alternatively, a joint embedding space -- might reduce that variance
and thus improve interpretability of the results.

\section*{Ethical Considerations}
In our work, we do not develop or annotate corpora. We further do not
collect data or propose new NLP tasks. Therefore, our work does not
contribute potential biases originating from annotator or data
selection.  Instead, our goal is to understand biases better and
contribute to a more fair emotion classification. We do not
investigate how topic bias might cause harm in downstream
applications.

Still, our topic analysis might be limited, for instance by the topic
modeler chosen for the analysis and by the datasets that we
studied. In real-world data applications, another topic modeling
approach might be required. It is important to note that we do not
make any statements which topics might have a negative impact
on members of a society.

In general, emotion classifiers have a high potential to cause harm by
making wrong predictions. Until the performance is on a higher, more
reliable level and the effects of biases and other confounding
variables are better understood, they should always be applied with
caution. We propose that the analyses acquired with automatic emotion
analysis methods should never be related to individuals. Instead,
analysis should only be performed on an aggregated level.

\bibliography{lit}

\clearpage

\appendix

\section{Implementation Details}
\label{app:implementation}
\paragraph{Emotion/Appraisal Classifier.}
Following state-of-the-art approaches to emotion and appraisal
classification \citep[\citealp{demszky-etal-2020}][]{Troiano2019}, we
fine-tune \roberta \citep{liu2019roberta} as implemented in the
Huggingface library \citep{wolf-etal-2020-transformers} on each
corpus.  For the classification, the output from the transformer
layers is pooled and passed through a fully-connected dense layer (768
units).  We apply ReLU activation \citep{agarap2019deep} and a dropout
of 0.5 and a consecutive classification layer using softmax activation
and binary cross-entropy loss for single-class classification (for
\isear, \tales, and emotions in \crowdenv).  For the multi-class
classification task (\ssec, \appreddit, \enisear and appraisals in
\crowdenv), we apply a sigmoid activation and categorical
cross-entropy loss instead.
The learning rate is set to $5 \times 10^{-5}$ across all experiments;
the batch size is 16.  We train each classifier for a maximum of 5
epochs and apply early stopping based on the validation accuracy
(stops after two consecutive epochs without improvement).  As
optimizer, AdamW \citep{loshchilov2019decoupled} is applied, weight
decay is set to $10^{-5}$.  Results are averaged over three different
runs for each classification task.

\paragraph{Topic Modeling.}
\bertopic consists of a pipeline of components for features
representation, dimensionality reduction, clustering and topic. We use
a pre-trained sentence embedding (all-MiniLM-L6-v2, as implemented in
Huggingface) for feature extraction, Accelerated Hierarchical Density
Clustering (HDBSCAN; \citealp{hdbscan}) as a clustering method,
Uniform Manifold Approximation (UMAP; \cite{mcinnes2020umap}) for
dimensionality reduction and tf-idf for retrieving the topics within
the clusters.  Although HDBSCAN does not require a pre-determined
number of topics, it can be tuned by setting hyperparameters for the
minimum cluster size and controlling the amount of outliers allowed
within a cluster.  We adapt these hyperparameters to each corpus
individually, depending on its size.

\paragraph{Word Removal.}
The list of topic words to be removed in each corpus consists of the
ten most representative words of each topic within the dataset.  The
most representative words, i.e., the top $k$ words per topic are
determined by the probability that \bertopic assigns to each word,
i.e., the word's probability to be assigned a certain topic label.
Therefore, $k$ is a hyperparameter determining the trade-off between
general classification performance and topic-influence: Increasing $k$
increases the potential impact of the de-biasing method (as less
topic-specific features are available to the classifier), but, at the
same time, decreases the general classification as less and less
features are available overall.  Further, by choosing a higher $k$,
more words which are less representative for a given topic are removed
as well, thus introducing noise to the experiment.  Here, $k$ is
set to 10. Setting$k = 3$ or $k = 5$ were considered as well, but did
not show a considerable change in performance compared to the
non-mitigated baseline classifier (\textsc{Bl}).  This hyperparameter
choice is further supported by the observation that the top $k$
representative words often comprise variations of the same word or
concept.  For example, in \isear, the ten most representative words
for the topic \textit{theft} consist of \say{theft}, \say{stealing},
\say{stole}, \say{thief}, \say{robbery}, \say{thieves}, \say{stolen},
\say{borrowed}, \say{robbers} and \say{cash}.  A higher $k$ thus
covers a broader range of morphological (\say{stealing}, \say{stole},
\say{stolen} and \say{thief}, \say{thieves}), as well as semantic
(\say{theft}, \say{robbery}) variation.  The chosen topic words are
not removed from the input, but substituted with \say{\dots}.  The
number of masked topic words per corpus is summarized in
\autoref{tab:removed_stopwords}.
\begin{table}
	\setlength{\tabcolsep}{4pt}
	\centering\footnotesize
	\begin{tabular}{l r r}
		\toprule
		& \# topics &\# masked topic words\\
		\cmidrule(r){1-1}
		\cmidrule(lr){2-2}
		\cmidrule(l){3-3}
		\isear & 10&100\\
		\ssec & 11&110\\
		\tales & 10&10\\
		\crowdenv & 8&80\\
		\appreddit & 10&100\\
		\enisear & 13&130\\
		\bottomrule
	\end{tabular}
	\caption{Number (\#) of topics and the resulting number of removed (i.e., masked) topic words.}
	\label{tab:removed_stopwords}
\end{table}

\paragraph{Gradient Reversal.}
The gradient reversal layer (GRL) is implemented as described by
\citet{ganin2015}, with the purpose of reversing the gradient (by
multiplying it with $-\lambda$) of the following layer during
backpropagation.  Since the layer has no trainable (nor non-trainable)
weights associated with it, the GRL has no effect during a forward
pass and acts as an identity transformation.  For the \ingrl and \crossgrl
experiments conducted here, the GRL is added into the standard
classifier architecture described above.  The emotion classifier is
coupled with an additional topic classification layer, equivalent to
the single-class emotion classification layer, with the task of
predicting the correct topic label $t_{i,c}$ for each instance.  The
topic classifier is connected via the GRL to the remaining layers of
the network, i.e., the pre-trained \roberta model as well as the
single dense layer.  Since the gradient is reversed, all weights in
the shared layer associated with the topic prediction task are
decreased.  A key factor in the implementation is the choice of
$\lambda$ as it regulates the impact of the GRL.  Again, choosing
$\lambda$ is a trade-off between overall classification performance
and de-biasing potency.  To determine an optimal value for $\lambda$,
standard emotion (or appraisal classifiers) are trained on each
individual corpus for $\lambda$ values of 0.1, 0.3, 0.5, 1 and 3.
Across corpora, a significant decrease in performance can be observed
for any $\lambda > 0.1$.  Therefore, $\lambda$ is set to 0.1 for all
gradient reversal experiments.

\section{Data}
\label{app:data}
Besides for their widespread use, the corpora are specifically selected for their variety in domain and text style.
As bias in general and topic bias in particular is closely related to the respective dataset's domain, annotation and sampling methods of a dataset, the following overview puts emphasis on these aspects.
We provide a detailed description of the datasets used in this investigation, emphasizing on each dataset's domain, annotation and sampling method. General corpus statistics are further provided in \autoref{tab:data_emo_annot}.

\begin{table*}
  \centering\footnotesize
  \begin{tabularx}{\textwidth}{X r r r r}
    \toprule
    Corpus
    & Size
    & Annotation
    & Domain
    & Class. Setting\\
    \cmidrule(r){1-1}\cmidrule(lr){2-2}\cmidrule(lr){3-3}\cmidrule(lr){4-4}\cmidrule(l){5-5}
    \isear
    & 7666
    & \joy, \sadness, \anger, \fear, \disgust, \shame, \guilt
    & event descr.
    & single \\
    \ssec
    & 4870
    & Plutchik
    & tweets
    & multi \\
    \tales
    & 10339
    & Ekman + \noemotion
    & fairy tales
    & single\\
    \textsc{crowd-enVENT}
    & 6600
    & Ekman + \shame, \pride, \textit{bored.}, \textit{rel.}, \trust, \guilt, \textit{no}
    & event descr.
    & single
    \\
    &
    & 21 appraisal dimensions
    &
    & multi\\
    \appreddit
    & 780
    & \textit{unexp.}, \textit{consist.}, \textit{cert.}, \textit{cntrl.}, \textit{resp.}
    & reddit posts
    & multi \\
    \enisear
    & 1001
    & \joy, \sadness, \anger, \fear, \disgust, \shame, \guilt
    & event descr.
    & single
    \\
    &
    & \textit{attent.}, \textit{cntrl.}, \textit{circum.}, \textit{resp.}, \textit{pleasant.}, \textit{effrt.}, \textit{cert.}
    &
    & multi \\
    \bottomrule
  \end{tabularx}
  \caption{Corpus overview. Emotion/appraisal statistics for
    \crowdenv/\enisear are reported separately.}
  \label{tab:data_stats}
\end{table*}

\subsection{Corpora}
\paragraph{\isear.}
The \isear corpus \citep{scherer-wallbott1994} consists of 7,665
sentences which were sampled in an in-lab setting: Participants were
presented with an emotion label and asked to report an event that
elicited that particular emotion in them.  Each event description is
labeled with a single emotion from a set of eight (Ekman's basic
emotions plus \textit{shame} and \textit{guilt}).  Since participants
were free to report any event that elicited one of the given emotions,
they were also free in their choice of topic.  However, since
participants were asked to report events specific to certain emotions,
sample bias could have been introduced to the corpus (under the
assumption that there are prototypical events for certain emotions).

\paragraph{\enisear.}
The corpus consist of 1001 event descriptions that were originally
compiled by \cite{Troiano2019} as a complement to \isear. The event
descriptions were sampled analogous to \isear, but in a crowd-sourcing
setup (annotated for \joy, \sadness, \anger, \fear, \disgust, \shame and \guilt).
Here, \enisear also refers to the appraisal annotations which were added to the
corpus by \citet{Hofmann2020}: \textit{Attention}, \certainty,
\effort, \pleasantness, \responsibility and \control.  These additional annotations
were provided by expert annotators.

\paragraph{\ssec.}
The Stance Sentiment Emotion Corpus \citep{Schuff2017} consists of
4,868 Twitter posts.  The original data stems from
\citet{mohammad-etal-2016-semeval} which \citet{Schuff2017}
re-annotate for Plutchik's eight basic emotions.  The annotations are
conducted by trained expert annotators.  Since the original dataset
by \citet{mohammad-etal-2016-semeval} was developed for stance
detection, the instances were sampled using keywords (i.e., hashtags)
that contain a particular stance in favor (e.g.,
\say{\#Hillary4President}) or against an entity (\say{\#HillNo}).
This type of keyword-based data sampling has been found to exhibit
topic bias in related studies, e.g., on datasets of abusive language
\citep{wiegand2019}.

\paragraph{\tales.}
The \tales corpus \citep{Alm2005} features 15,302 sentences from
different fairytales.  Sentences are labeled by experts with one of
Ekman's basic emotions (\textit{surprise} is split into
\textit{negative} and \textit{positive surprise}).  Emotions are
annotated from the perspective of the respective character.

\paragraph{\crowdenvlong.}
Analogous to \textsc{enISEAR}, the \crowdenvlong corpus
\citep{troiano-etal2023} consists of 6600 crowd-sourced, self-reported
event descriptions.  Each description is annotated for 21 appraisal
dimensions\footnote{ Suddenness, familiarity, event predictability,
  pleasantness, unpleasantness, goal relevance, own responsibility,
  others' responsibility, situational responsibility, anticipation of
  consequences, goal support, urgency, own control, others' control,
  situational control, acceptance of consequences, clash with internal
  standards and ideals, violation of (external) norms and laws, not
  consider, attention, effort.  }, each rated on a scale between 1 and
5, as well as for emotions (Ekman's 6 basic emotions, plus
\textit{shame}, \textit{pride}, \textit{boredom}, \textit{relief},
\textit{trust}, \shame, \guilt and \noemotion).  Participants were
free in their choice of topic, but the priming with an emotion label
might influence the topic distribution (see \isear).  In order to avoid
oversampling descriptions of prototypical events,
\citeauthor{troiano-etal2023} apply a diversification method to foster
more diverse event descriptions.
The corpus additionally features crowd-sourced re-annotations of the
event descriptions to investigate differences between the reader's
and writer's assessment of emotions and appraisals. However, these
are not used here.

\paragraph{\appreddit.}
The \appreddit corpus \citep{Stranisci2022} is annotated with
appraisal dimensions.  It comprises 780 reddit posts, where each posts
contains at least one event description (1,091 events overall).  The
five appraisal labels (\textit{certainty}, \textit{consistency},
\textit{control}, \textit{unexpectedness}, \textit{responsibility})
are based on \cite{roseman1991} and annotated by experts.  The posts
are sampled exclusively from a limited set of subreddits, mostly
connotated with negative sentiment (Anger, offmychest, helpmecope
anxiety, i.a.). This sampling procedure might introduce bias to the
dataset.

\subsection{Aggregated Annotation Scheme}
As depicted above, the corpora differ in their annotation schemes.  In
order to provide a more comparable analysis, the individual
annotations are mapped onto an inter-corpora annotation scheme.  For
emotions, \anger, \disgust, \fear, \joy, \sadness, \shame, \surprise,
\noemotion and \other are considered.  This subset of emotion labels
is based on basic emotions \cite{ekman1999}. Beyond
\citeauthor{ekman1999}'s six emotions, the list accounts for other
labels that frequently occur (see \autoref{tab:data_emo_annot} for an
overview).  The same procedure is applied to appraisal
labels. However, approaches to appraisal classification are even more
diverse in annotation than emotion datasets. To account for this
variation, the inter-corpora labelset consists of 11 appraisal
dimensions (suddenness, pleasantness, self control, chance control,
self responsibility, other responsibility, goal support, predict
consequences, attention, effort), however, only a subset of six labels
is shared across two of the three corpora annotated with appraisals,
while only two labels can be mapped to all three corpora (summarized
in \autoref{tab:data_appr_annot}).

\section{Other Emotion--Topic Relations}
\label{app:pmi}
\begin{figure}[h]
\includegraphics[scale=0.3]{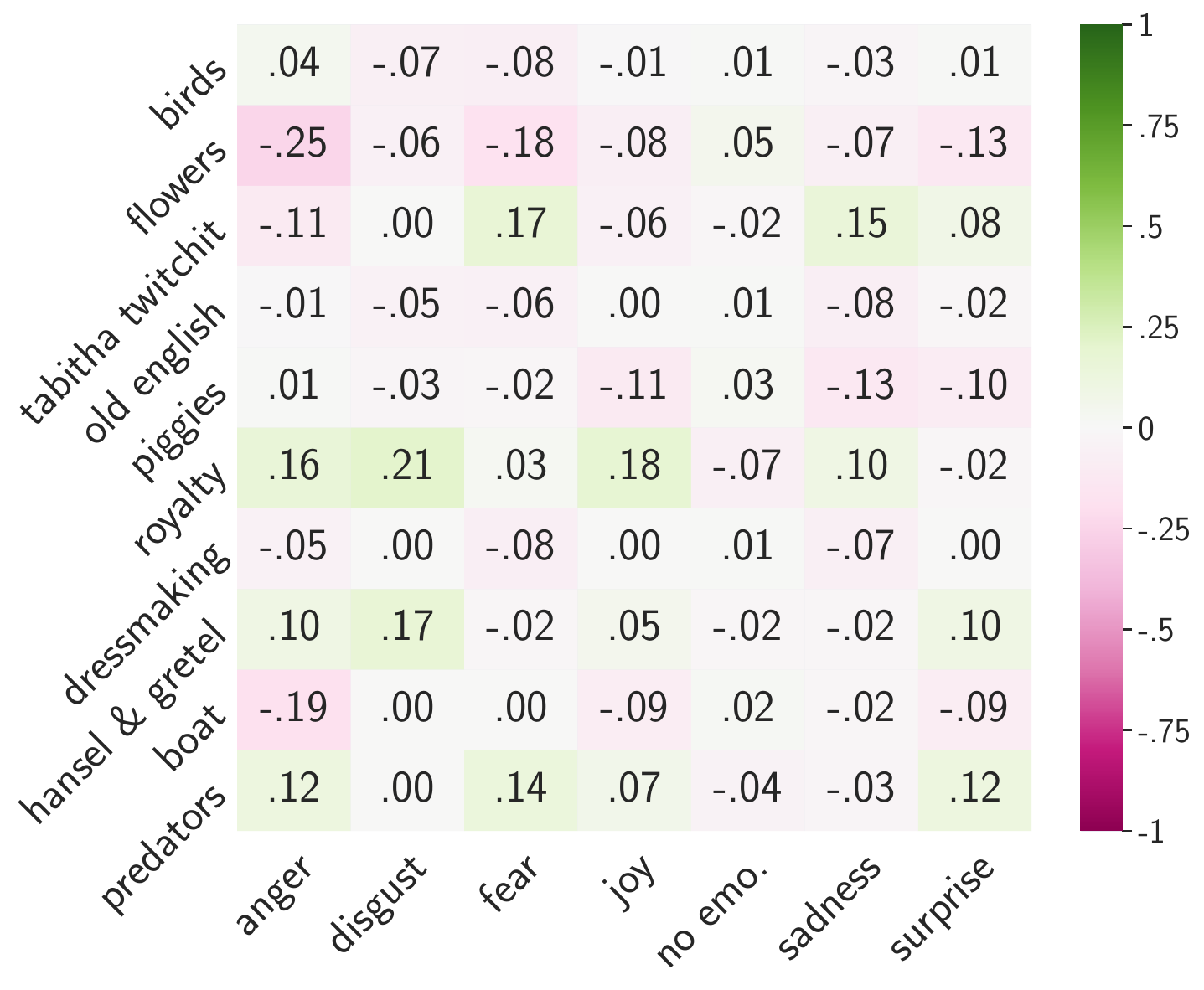}
\caption{Normalized pointwise mutual information between topics and emotion annotations in \tales.}
\label{fig:pmi_tales}
\end{figure}

\begin{figure}[h]
\includegraphics[scale=0.3]{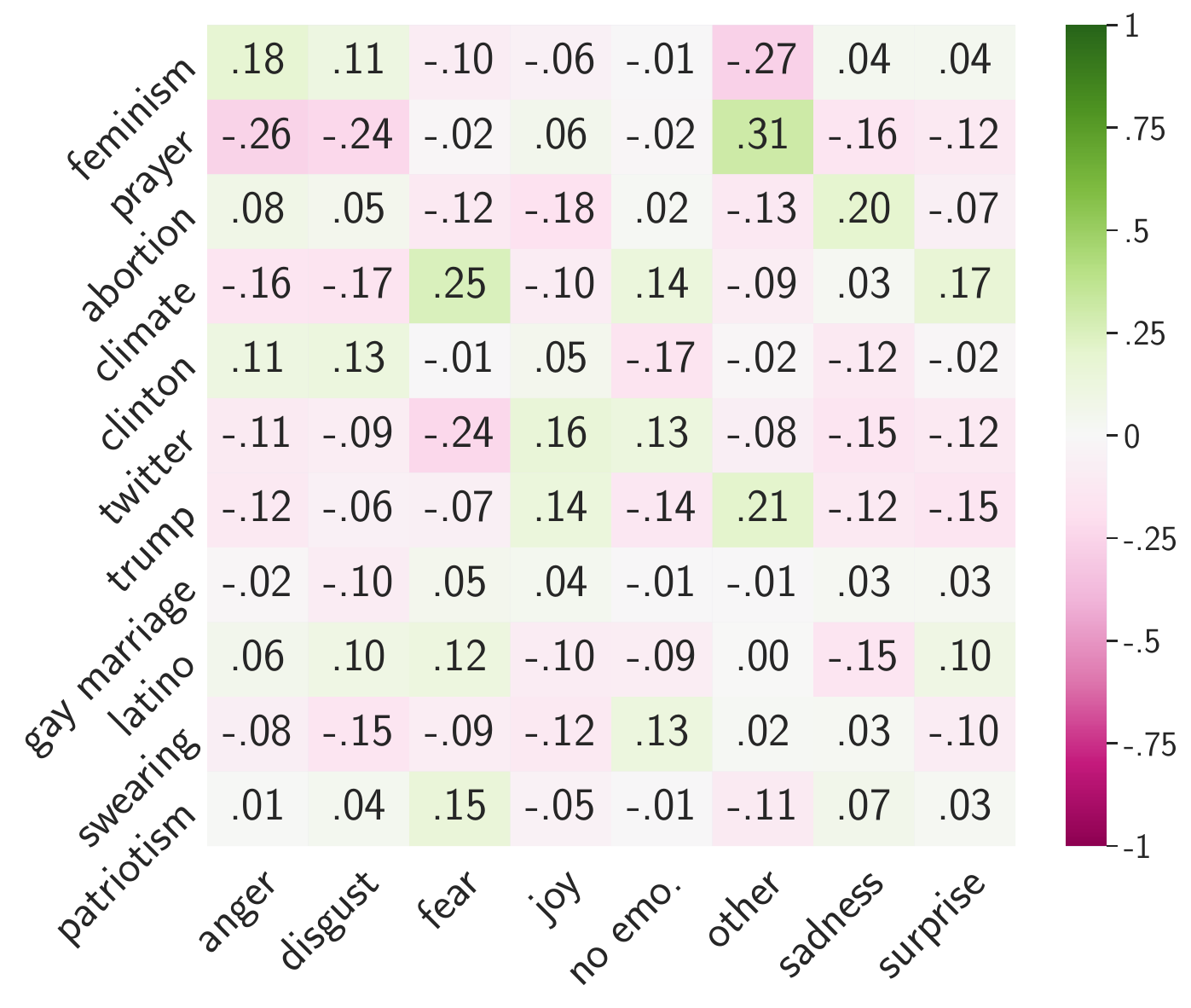}
\caption{Normalized pointwise mutual information between topics and emotion annotations in \ssec.}
\label{fig:pmi_ssec}
\end{figure}

Figures~\ref{fig:pmi_tales} and \ref{fig:pmi_ssec} show the results
for topic--emotion associations for the \tales and the \ssec corpora,
analogously to the other resources in
Section~\ref{sec:results}.\\

\clearpage
\onecolumn

\begingroup
  \centering
  \begin{tabularx}{\textwidth}{X r r r r r r r r r}
    \toprule
    Corpus
    & \multicolumn{1}{c}{A} %Anger
    & \multicolumn{1}{c}{D} % Disgust
    & \multicolumn{1}{c}{F} % Fear
    & \multicolumn{1}{c}{J} % Joy
    & \multicolumn{1}{c}{Sa} % Sadness
    & \multicolumn{1}{c}{Sh} % Shame
    & \multicolumn{1}{c}{Su} %Surprise
    & \multicolumn{1}{c}{No} %No-emo
    & \multicolumn{1}{c}{O} % Other
    \\
    \cmidrule(r){1-1}\cmidrule(lr){2-2}\cmidrule(lr){3-3}\cmidrule(lr){4-4}
    \cmidrule(r){5-5}\cmidrule(lr){6-6}\cmidrule(lr){7-7}\cmidrule(lr){8-8}
    \cmidrule(r){9-9}\cmidrule(lr){10-10}
    { \small \crowdenv} % Crowd-enVENT
    & {\small 550}% anger
    & {\small 550}% disgust
    & {\small 550}% fear
    & {\small 550}% joy
    & {\small 550}% sadness
    & {\small 550}*% shame
    & {\small 550}% surprise
    & {\small 550}% no-emotion
    & {\small 2,200*}% other
    \\
    {\small \isear} % ISEAR
    & {\small 1,096}% anger
    & {\small 1,096}% disgust
    & {\small 1,095}% fear
    & {\small 1,094}% joy
    & {\small 1,096}% sadness
    & {\small 2,189*}% shame
    & {\small$-$}% surprise
    & {\small$-$}% no-emotion
    & {\small$-$}% other
     \\
    {\small \enisear} % enISEAR
    & {\small 143}% anger
    & {\small 143}% disgust
    & {\small 143}% fear
    & {\small 143}% joy
    & {\small 143}% sadness
    & {\small 286*}% shame
    & {\small$-$}% surprise
    & {\small$-$}% no-emotion
    & {\small$-$}% other
    \\
    {\small \ssec} % SSEC
    & {\small 1388}% anger
    & {\small 440}% disgust
    & {\small 274}% fear
    & {\small 815}% joy
    & {\small 414}% sadness
    & {\small$-$}% shame
    & {\small 177}% surprise
    & {\small 1552}% no-emotion
    & {\small 1077*}% other
    \\
    {\small \tales} % Tales
    & {\small 302}% anger
    & {\small 40}% disgust
    & {\small 251}% fear
    & {\small 579}% joy
    & {\small 340}% sadness
    & {\small$-$}% shame
    & {\small 144}% surprise
    & {\small 8,683}% no-emotion
    & {\small $-$}% other
    \\
    \bottomrule
  \end{tabularx}
  \captionof{table}{\label{tab:data_emo_annot} Number of instances of each emotion class (after mapping; the asterisk (*) indicates that this class includes mapped labels, i.e., combining multiple classes into one aggregated, but not simple one-to-one mapping of equivalent labels (happiness $\rightarrow$ joy).}
  \endgroup
  \vspace{1cm}

\begingroup
  \centering\small
  \begin{tabularx}{\textwidth}{X r r r r r r r r r r r }
    \toprule
    Corpus
    & \parbox[t]{2mm}{\rotatebox[origin=lB]{60}{Attention}}
    & \parbox[t]{2mm}{\rotatebox[origin=lB]{60}{Pleasantness}}
    & \parbox[t]{2mm}{\rotatebox[origin=lB]{60}{Suddenness}}
    & \parbox[t]{2mm}{\rotatebox[origin=lB]{60}{Self Control}}
    & \parbox[t]{2mm}{\rotatebox[origin=lB]{60}{Chance Control}}
    & \parbox[t]{2mm}{\rotatebox[origin=lB]{60}{Self Responsibility}}
    & \parbox[t]{2mm}{\rotatebox[origin=lB]{60}{Other Responsibility}}
    & \parbox[t]{2mm}{\rotatebox[origin=lB]{60}{Predict Consequences}}
    & \parbox[t]{2mm}{\rotatebox[origin=lB]{60}{Goal Support}}
    & \parbox[t]{2mm}{\rotatebox[origin=lB]{60}{Effort}}
    & \parbox[t]{2mm}{\rotatebox[origin=lB]{60}{Other}}
    \\
    \cmidrule(r){1-1}
    \cmidrule{2-12}
    { \small \appreddit} % APPreddit
    & {\small $-$}% attention
    & {\small $-$}% pleasantness
    & {\small 307}% suddenness
    & {\small 307}% self control
    & {\small $-$}% chance control
    & {\small 400}% self responsibility
    & {\small 457}% other responsibility
    & {\small 748}% predict consequences
    & {\small 312}% goal support
    & {\small $-$}% effort
    & {\small $-$}% other
    \\
    {\small \crowdenv} % Crowd-enVENT
    & {\small 4125}% attention
    & {\small 2261}% pleasantness
    & {\small 3128}% suddenness
    & {\small 2142}% self control
    & {\small 1514}% chance control
    & {\small 2597}% self responsibility
    & {\small 3396}% other responsibility
    & {\small 2841}% predict consequences
    & {\small 2281}% goal support
    & {\small 3210}% effort
    & {\small 6527*}% other
    \\
    {\small \enisear} % enISEAR
    & {\small 673}% attention
    & {\small 149}% pleasantness
    & {\small $-$}% suddenness
    & {\small 228}% self control
    & {\small 240}% chance control
    & {\small 377}% self responsibility
    & {\small $-$}% other responsibility
    & {\small 761}% predict consequences
    & {\small $-$}% goal support
    & {\small 400}% effort
    & {\small $-$}% other
    \\
    \bottomrule
  \end{tabularx}
  \captionof{table}{\label{tab:data_appr_annot} Number of instances of each appraisal class (after mapping; the asterisk (*) indicates that this class includes mapped labels, either by simple one-to-one mapping (happiness $\rightarrow$ joy), or by combining multiple classes into one aggregated).}

\endgroup

\end{document}